%% file: main.tex
\def\year{2021}\relax
\documentclass[letterpaper]{article} 
\usepackage{aaai21}  
\usepackage{times}  
\usepackage{helvet} 
\usepackage{courier}  
\usepackage[hyphens]{url}  
\usepackage{graphicx} 
\usepackage{graphicx}  
\usepackage{natbib}  
\usepackage{caption} 
\usepackage{booktabs}
\usepackage{pifont}
\newcommand{\cmark}{\ding{51}}
\newcommand{\xmark}{\ding{55}}
\include{def}

\usepackage{color}

\title{Motion-blurred Video Interpolation and Extrapolation }
\author {

        Dawit Mureja Argaw \hspace{0.4cm}
        Junsik Kim \hspace{0.4cm}
        Francois Rameau \hspace{0.4cm}
        In So Kweon\\
}
\affiliations {
    Dept. of Electrical Engineering, KAIST, Daejeon, Korea \\
    dawitmureja@kaist.ac.kr, \{mibastro, rameau.fr\}@gmail.com, iskweon77@kaist.ac.kr
}
\begin{document}

\maketitle
\begin{abstract}
     Abrupt motion of camera or objects in a scene result in a blurry video, and therefore recovering high quality video requires two types of enhancements: visual enhancement and temporal upsampling. A broad range of research attempted to recover clean frames from blurred image sequences or temporally upsample frames by interpolation, yet there are very limited studies handling both problems jointly. In this work, we present a novel framework for deblurring, interpolating and extrapolating sharp frames from a motion-blurred video in an end-to-end manner. We design our framework by first learning the pixel-level motion that caused the blur from the given inputs via optical flow estimation and then predict multiple clean frames by warping the decoded features with the estimated flows. To ensure temporal coherence across predicted frames and address potential temporal ambiguity, we propose a simple, yet effective flow-based rule. The effectiveness and favorability of our approach are highlighted through extensive qualitative and quantitative evaluations on motion-blurred datasets from high speed videos.

\end{abstract}
\section{Introduction}
Video frame interpolation aims at predicting one or more intermediate frames from given input frames for high frame-rate conversion. Existing frame interpolation approaches can be broadly categorized into flow-based~\cite{mahajan2009moving,zitnick2004high,liu2017video}, kernel-based~\cite{niklaus2017video1,niklaus2017video2,Lee_2020_CVPR} and a fusion of the two~\cite{bao2019memc,DAIN}. Intermediate frames are interpolated either by directly warping the input frames with estimated optical flows (motion kernels) or using a trainable frame synthesis network. Extending these approaches for motion-blurred videos, however, is not a trivial process. Blurry video is a result of abrupt motions and long exposure time. As a result, contents in the video are degraded by motion blur and the gap between frames is relatively large compared to normal videos. This makes the computation of optical flow or motion kernel very challenging resulting in a subpar network performance (see \Tref{tbl:interp}).

There have been limited studies on joint deblurring and interpolation of a motion-blurred video. A na\"ive approach to the task at hand would be to cascade deblurring and interpolation methods interchangeably. With the recent progress in motion deblurring, several deep network based single image \cite{Nah_2017_CVPR,Zhang_2018_CVPR,tao2018srndeblur} and video \cite{hyun2015generalized,su2017deep,nah2019recurrent} deblurring approaches have been proposed. Given a blurry video, deploying deblurring frameworks followed by interpolation methods to predict sharp intermediate frames is not optimal since deblurring artifacts would propagate across the interpolated frames. Similarly, interpolation followed by deblurring would result in the propagation of interpolation artifacts caused by imprecise optical flow (motion kernel) predictions.  

Recent video restoration works \cite{Jin_2018_CVPR,purohit2019bringing} attempted to extract multiple clean frames from a single motion-blurred image. Applying these works for blurry video interpolation (and extrapolation) by successively feeding blurry inputs, however, is problematic due to temporal ambiguity. A closely related work by Jin~\etal \cite{Jin_2019_CVPR} jointly optimized deblurring and interpolation networks to predict clean frames from four blurry inputs. A concurrent work by Shen \etal \cite{shen2020blurry} proposed an interpolation module that outputs a single sharp frame from two blurry inputs. More frames are generated by applying the interpolation module on the predicted sharp frames in a recurrent manner.

In this work, we propose a novel framework to \textit{interpolate} and \textit{extrapolate} multiple sharp frames from two blurry inputs. Inspired by the fact that motion-blurred image is a temporal aggregation of several latent frames during the exposure time of a camera, we exploit the input blurs as motion cues that can be leveraged to better infer and account for inter-frame motion. This is achieved by decoding latent frame features via optical flow estimation. We also design a flow-based rule to address temporal ambiguity and predict frames by warping the decoded features with the estimated flows. Unlike previous works \cite{Jin_2019_CVPR,shen2020blurry} that implicitly follow a deblurring $\rightarrow$ interpolation pipeline to predict intermediate frames between the deblurred middle latent frames, we adopt a motion-based approach to interpolate and extrapolate the entire latent frame sequence directly from given inputs in a temporally coherent manner.

We evaluated the proposed approach qualitatively and quantitatively on real image blur datasets generated from high speed videos \cite{Nah_2017_CVPR,Jin_2019_CVPR}. We also comprehensively analyzed our work in connection to various related approaches on motion-blurred video interpolation, extrapolation and deblurring tasks to highlight the effectiveness and favourability of our approach. Moreover, we provide generalization experiments on real motion-blurred videos from \cite{Nah_2019_CVPR_Workshops_REDS,su2017deep}. In short, our contributions are: (1). We present a novel and intuitive framework for motion-blurred video interpolation and extrapolation (2). We propose a simple, yet effective, flow-based rule to address potential ambiguity and to restore latent frames in a temporally coherent manner
(3). We extensively analyze our approach in relation to previous works and obtain a favourable performance.
(4). We showcase the applicability of our model for related tasks such as video deblurring and optical flow estimation from motion-blurred inputs
(5). We provide detailed ablation study on different network components to shed a light on the network design choice.

\section{Methodology}
\paragraph{Background.}
Motion-blurred image is a temporal average of multiple latent frames captured due to a sudden camera shake or dynamic motion of objects in a scene during the exposure time of a camera.

\begin{equation}
    B_t = \int_{t}^{t+e} L(\tau) d\tau ,
\end{equation}
where $L(\tau)$ is a latent frame at time $\tau$, $e$ is the exposure time and $B$ is the resulting motion-blurred image. As manually capturing a large blur dataset is a daunting task, a common practice in computer vision research is to synthesize a motion-blurred image by averaging consecutive frames in a high speed video \cite{DeblurGAN,Nah_2017_CVPR,tao2018srndeblur,su2017deep,Jin_2018_CVPR,Jin_2019_CVPR,shen2020blurry}.

\begin{equation}
    B_t = \frac{1}{N}\sum_{i = t-N/2}^{t+N/2} I_{i} ,
\end{equation}
where $N$ is the number of frames to average and $I_i$ is a clean frame in high speed video at time index $i$.

Given a blurred input $B_t$, image and video deblurring approaches \cite{DeblurGAN,Nah_2017_CVPR,tao2018srndeblur,su2017deep} recover the middle latent frame $I_{t}$. Recent works \cite{Jin_2018_CVPR,purohit2019bringing} attempted to restore the entire latent frame sequence from a single motion-blurred input, \ie $\{I_{t-N/2},\ldots,I_{t+N/2}\}$. However, these works suffer from a highly ill-posed problem known as \emph{temporal ambiguity}. Without the help of external sensors such as IMU or other clues on the camera motion, it is not possible to predict the correct temporal direction from a single motion-blurred input as averaging does not preserve temporal order, \ie both backward and forward averaging of sequential latent frames result in the same blurred image. Hence, deploying such methods \cite{Jin_2018_CVPR,purohit2019bringing} for motion-blurred video interpolation by successively feeding blurry frames is problematic as temporal coherence in the interpolated frames cannot be guaranteed.

Given two (or more) blurry frames, motion-blurred video interpolation aims at predicting sharp intermediate frames.
\begin{equation}
    B_{t_0} = \frac{1}{N}\sum_{i = t_{0}-N/2}^{t_{0}+N/2} I_{i} \hspace{2mm} \ldots \hspace{2mm}  B_{t_n} = \frac{1}{N}\sum_{i = t_{n}-N/2}^{t_{n}+N/2} I_{i}
\end{equation}
, where $t_n \le t_{n-1} + N$. 
Recent work by Jin \etal \cite{Jin_2019_CVPR} attempted to extract clean frames from four blurry inputs $\{B_{t_0},\ldots,B_{t_4}\}$ by first recovering the corresponding middle latent frames, \ie $\{I_{t_{0}},\ldots,I_{t_{4}}\}$ using a deblurring network and then generating more intermediate frames between the recovered latent frames using an interpolation network. Compared to a naive approach of cascading deblurring and interpolation frameworks, their method is optimized in an end-to-end manner. A concurrent work by Shen \etal~\cite{shen2020blurry} proposed a pyramidal recurrent framework without explicit deblurring followed by interpolation. Given two blurry frames $B_{t_0}$ and $B_{t_1}$, their approach directly outputs a sharp intermediate frame $I_{t_{0}+\Delta}$, where $\Delta \approx (t_1 - t_0)/2$. They addressed joint blur reduction and frame rate up-conversion by consecutively inputting blurry frame pairs and recursively applying the same procedure on the predicted sharp frames.

\begin{figure*}[!t]
\includegraphics[width=1.0\linewidth,trim={1.22cm 2.0cm 0.7cm 1.3cm},clip]{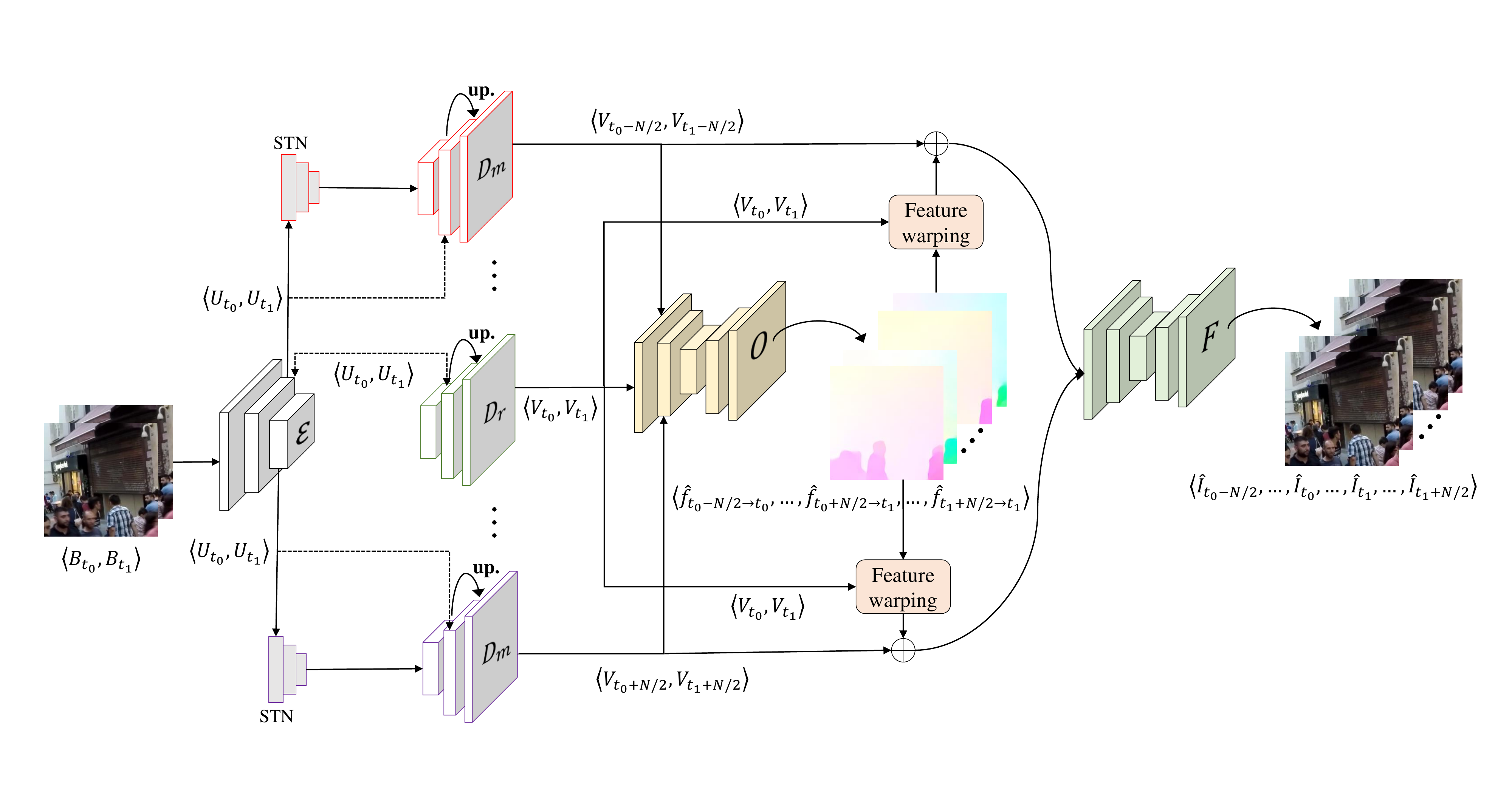} 
\caption{Overview of the proposed framework. First, we encode features from the given blurry inputs. Then, we decode latent frame features from the encoded features using global and local motion decoders that are supervised via optical flow estimation. Finally, we reconstruct multiple sharp frames in a bottom-up manner by warping the decoded features with the estimated flows.}
\label{fig:model}
\vspace{-2mm}
\end{figure*}

\paragraph{Problem formulation.}
In this work, we tackle the problem of interpolating and extrapolating multiple clean frames from blurry inputs in a single stage. Given two motion-blurred inputs $B_{t_0}$ and $B_{t_1}$, we aim to recover all latent frames, \ie $\{I_{t_{0}-N/2},\ldots,I_{t_{o}+N/2},I_{t_{1}-N/2},\ldots,I_{t_{1}+N/2}\}$. For brevity, we refer to the middle latent frames ($I_{t_{0}}$ and $I_{t_{1}}$) as \textit{reference} frames. We propose a novel method to interpolate the intermediate latent frames  between the reference frames ($\{I_{t_{0}},\ldots,I_{t_{1}}\}$) and to extrapolate the past ($\{I_{t_{0}-N/2},\ldots,I_{t_{0}-1}\}$) and future ($\{I_{t_{1}+1},\ldots,I_{t_{1}+N/2}\}$) latent frames in one forward pass. We design our algorithm as follows. First, we encode and decode latent features by learning the pixel-level motion that occurred between the latent frames via optical-flow estimation. Second, we establish a simple, yet effective, flow-based rule to address potential temporal ambiguity. Third, we interpolate multiple clean frames by warping the decoded reference features with the estimated optical flows (see \Fref{fig:model}). 

Our work is different from previous works \cite{Jin_2019_CVPR, shen2020blurry} in the following aspects: 1. Our approach interpolates multiple clean frames directly from two blurry inputs in a single stage while previous works recursively apply an interpolation module on the predicted clean frames. 2. We adopt a motion-based approach to interpolate intermediate latent frames rather than predicting frames in a generic manner, thereby showing that our approach is relatively robust in handling large motions.
3. Previous works only focus on interpolation \ie deblurring the reference latent frames and interpolating intermediate frames between them. They ignore the other latent frames in order not to deal with temporal ambiguity, and hence, their work can not be extended for extrapolation task. By contrast, we interpolate and extrapolate latent frames in a temporally coherent manner by addressing temporal ambiguity with the proposed motion-based approach. 

\section{Proposed approach}
\paragraph{Feature encoding and decoding.}
Given two blurry inputs $B_{t_0}$ and $B_{t_1}$, an encoder network $\calE$ is used to extract feature representations of each input at different levels of abstractions (\Eref{eqn:encoding}). The encoder $\calE$ is a feed-forward CNN with five convolutional blocks each with two layers of convolutions of kernel size $3\times 3$ and stride size of 2 and 1, respectively.
\begin{equation}
    \{U_{t_0}^{l}\}_{l=1}^{K} = \calE(B_{t_0}) \hspace{0.5cm}\{U_{t_1}^{l}\}_{l=1}^{K} = \calE(B_{t_1})
    \label{eqn:encoding}
\end{equation}
, where $U_{t_0}^l$ is an encoded feature of $B_{t_0}$ at level $l$ and $K$ (fixed to 6 in our experiments) is the number of levels (scales) in the feature pyramid.

The encoded features are then decoded into latent frame features as shown in \Fref{fig:model}. Reference (middle) features are directly decoded by successively upsampling the encoded features using layers of transposed convolution of kernel size $4\times 4$ and a stride size of 2. 
A reference latent feature decoder $\calD_r$ inputs the $\times2$ upsampled decoded reference feature from level $l+1$ and the corresponding encoded feature concatenated channel-wise as shown in \Eref{eqn:mid_decoding}.

\begin{equation}
    \{V_{t_0}^l,V_{t_1}^l\} = \calD^l_r\big(\mathrm{\textbf{up.}}\{V_{t_0}^{l+1},V_{t_1}^{l+1}\} \oplus \{U_{t_0}^{l},U_{t_1}^{l} \}\big)
    \label{eqn:mid_decoding}
\end{equation}
, where  $\mathrm{\textbf{up.}}$ stands for upsampling, $\oplus$ is for channel-wise concatenation, $V_{t_0}$ denotes the decoded reference latent feature. The other (non-middle) features are decoded by inferring the blur motion from the encoded features. In order to learn the global motion of the other latent frames with respect to the reference latent frame, we used spatial transformer networks (STNs) \cite{jaderberg2015spatial}. Given an encoded feature, STN estimates an affine transformation parameter $\theta_{[R|T]}$ to spatially transform the input feature. As STN is limited to capturing only global motion, in order to compensate for the apparent local motions, we further refine the transformed feature using a motion decoder. A motion decoder $\calD_m$ inputs the globally transformed feature along with the encoded feature (via skip connection shown in \Fref{fig:model} in dotted lines) and the $\times2$ upscaled non-middle latent feature from level $l+1$, and outputs a decoded a non-middle latent feature at level $l$ as follows,
\begin{equation}
    {V_{s_0}^{l}} = \calD_{m}^{l}\big(\mathrm{STN}^{l}_{s_0}\{U_{t_0}^{l}\} \oplus \mathrm{\textbf{up.}}\{V_{s_0}^{l+1}\} \oplus U_{t_0}^{l}\big)
    \label{eqn:nonmid_decoding1}
    \end{equation}
\begin{equation}
    {V_{s_1}^{l}} = \calD_{m}^{l}\big(\mathrm{STN}^{l}_{s_1}\{U_{t_1}^{l}\} \oplus \mathrm{\textbf{up.}}\{V_{s_1}^{l+1}\} \oplus U_{t_1}^{l}\big)
    \label{eqn:nonmid_decoding2}
    \end{equation}

, where $s_0 = \{t_0-N/2,\ldots,t_{0}-1,t_{0}+1,\ldots,t_0+N/2\}$ and $s_1 = \{t_1-N/2,\ldots,t_{1}-1,t_{1}+1,\ldots,t_1+N/2\}$.
\paragraph{Optical flow estimation.}
Our network learns to decode latent frame features from the blurry inputs via optical flow estimation, \ie the optical flow between the latent frames is computed using the respective decoded features. For instance, to estimate the optical flow between $I_{t_{0}-N/2}$ and $I_{t_0}$, the corresponding decoded features $\{V_{t_{0}-N/2}\}_{l=1}^{K}$ and  $\{V_{t_{0}}\}_{l=1}^{K}$ are used. The two sets of decoded features here are equivalent to the encoded features of two clean input images in standard optical flow estimation algorithms. We estimate flow in a coarse-to-fine manner mimicking the vanilla pipeline for optical flow estimation from two images \cite{Sun2018PWC-Net,fischer2015flownet,hui2018liteflownet,ranjan2017optical,IMKDB17}. Given two decoded features at feature level $l$ (\eg $V_{t_{0}-N/2}^{l}$ and $V_{t_0}^{l}$), a warping layer $\calW$ is used to back-warp the second feature $V_{t_0}^{l}$ (to the first feature $V_{t_{0}-N/2}^{l}$ ) with $\times2$ upsampled flow from level $l+1$ as shown in \Eref{eqn:warping}. A correlation layer $\calC$ \cite{fischer2015flownet,Sun2018PWC-Net} is then used to compute the matching cost (cost volume) between the first feature $V_{t_{0}-N/2}^{l}$ and the back-warped second feature $\hat{V}_{t_{0}}^{l}$. The optical flow $\hat{f}^l$ is estimated using an optical flow estimator network $\calO$ that inputs the cost volume, the first feature and the upsampled optical flow  concatenated channel-wise and outputs a flow (\Eref{eqn:flow_esti}). Following \cite{Sun2018PWC-Net}, we use a context network to refine the estimated full-scale flow.

\begin{equation}
    \hat{V}_{t_{0}}^{l} = \calW\big(V_{t_{0}-N/2}^{l}, \mathbf{up.}\{\hat{f}^{l+1}\}\big)
    \label{eqn:warping}
\end{equation}
\begin{equation}
    \hat{f}^l = \calO\big(\calC\{V_{{t_0}-N/2}^{l},\hat{V}_{{t_0}}^{l}\} \oplus V_{{t_0}-N/2}^{l} \oplus \mathbf{up.}\{\hat{f}^{l+1}\}\big)
    \label{eqn:flow_esti}
\end{equation}

\begin{figure}[!t]
\centering
\includegraphics[width=1.0\linewidth]{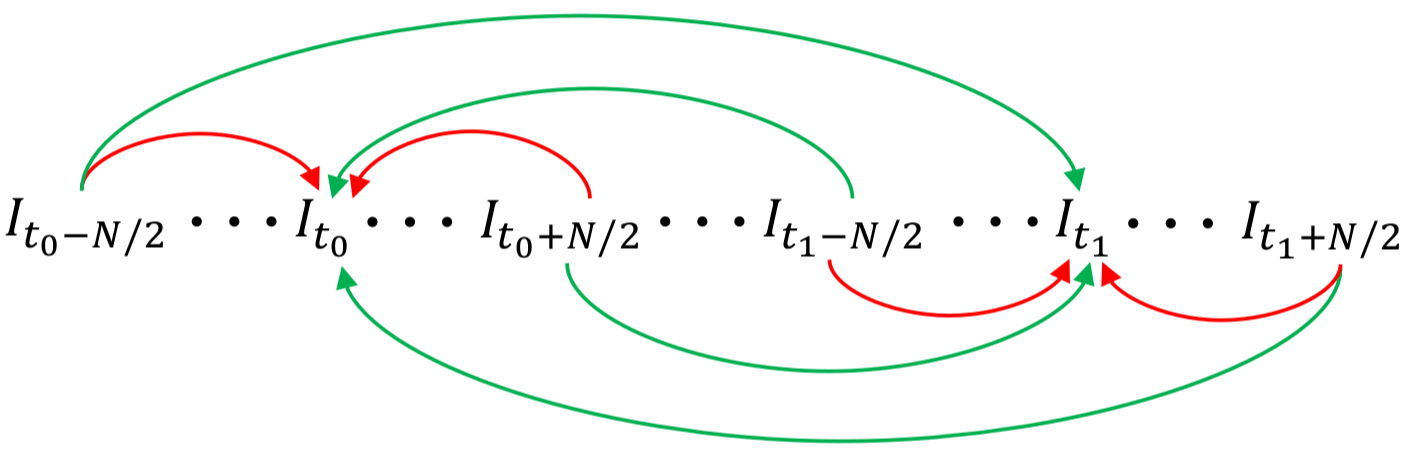}
\caption{Optical flow estimation between latent frames}
\vspace{-2.5mm}
\label{fig:flow}
\end{figure}

In the same manner, we predict multiple optical flows between latent frames (see \Fref{fig:flow}). Since the ground truth optical flow is not available to train the flow estimator, we used a pretrained FlowNet 2 \cite{IMKDB17} network to obtain pseudo-ground truth flows (between sharp latent frames) to supervise the optical flow estimation between the decoded features. The flow supervision via imperfect ground truth flows is further enhanced by the frame supervision as our network is trained in an end-to-end manner. 

\paragraph{Temporal ordering and ambiguity.} Estimating optical flow between decoded features is crucial for maintaining temporal coherence across the predicted frames. We estimate optical flow (shown in red in \Fref{fig:flow}) between the reference latent frame and non-middle latent frames within each blurry input, \ie $\{\hat{f}_{{t_0}-N/2\rightarrow{t_0}},\ldots,\hat{f}_{{t_0}+N/2\rightarrow{t_0}}\}$ and $\{\hat{f}_{{t_1}-N/2\rightarrow{t_1}},\ldots,\hat{f}_{{t_1}+N/2\rightarrow{t_1}}\}$. Constraining these flows enforces our model to learn motions in a symmetric manner with STNs and motion decoders close to the reference features decoding smaller motions, and those further from the reference features decoding larger motions. This in turn preserves temporal ordering within the decoded features of each blurry input avoiding random shuffling. However, correct temporal direction can not be still guaranteed as features can be decoded in a reverse order. To address this potential temporal ambiguity, we propose a simple, yet effective flow-based rule. We predict optical flow (shown in green in \Fref{fig:flow}) between the non-middle latent frames of the first input and the reference latent frame of the second input and vice versa, \ie $\{\hat{f}_{{t_0}-N/2\rightarrow{t_1}},\ldots,\hat{f}_{{t_0}+N/2\rightarrow{t_1}}\}$ and $\{\hat{f}_{{t_1}-N/2\rightarrow{t_0}},\ldots,\hat{f}_{{t_1}+N/2\rightarrow{t_0}}\}$. By constraining these flows via endpoint error supervision, we establish the following rules:

\textit{\underline{Rule 1}}. if $\|\hat{f}_{{t_0}-N/2\rightarrow{t_1}}\|$ $>$ $\|\hat{f}_{{t_0}+N/2\rightarrow{t_1}}\|$, it means that the features of $B_{t_0}$ are decoded in the correct order \ie $\{V_{{t_0}-N/2},\ldots,V_{t_0},\ldots,V_{{t_0}+N/2}\}$.

\textit{\underline{Rule 2}}. if $\|\hat{f}_{{t_0}-N/2\rightarrow{t_1}}\|$ $<$ $\|\hat{f}_{{t_0}+N/2\rightarrow{t_1}}\|$, it means that the features of $B_{t_0}$ are decoded in a reverse order \ie $\{V_{{t_0}+N/2},\ldots,V_{t_0},\ldots,V_{{t_0}-N/2}\}$ and hence, should be reversed to the correct order.

, where $\|\cdot\|$ denotes the magnitude of the flow. In a similar manner, we can use the optical flows $\hat{f}_{{t_1}-N/2\rightarrow{t_0}}$ and $\hat{f}_{{t_1}+N/2\rightarrow{t_0}}$ to ensure that the features of $B_{t_1}$ are decoded in the correct order. These rules need to be applied only on the four flows between the latent decoded features on the extrema ($V_{{t_0}-N/2},V_{{t_0}+N/2},V_{{t_1}-N/2},V_{{t_1}+N/2}$) and the reference decoded features, since temporal ordering within each input is maintained. Hence, the proposed flow-based rule can be used to interpolate and extrapolate larger number of frames with no additional computational cost.

\paragraph{Frame synthesis.} The decoded features and the estimated optical flows are then used to interpolate and extrapolate sharp frames from the blurry inputs. The reference latent frames are predicted at different spatial scales directly from the decoded reference features using a frame synthesis network $\calF_r$ (\Eref{eqn:mid_frame}). This is equivalent to deblurring each input frame except for the fact that we output deblurred middle frames at different scales. The other (non-middle) latent frames are predicted by back-warping the decoded reference features with the corresponding optical flows. For better reconstruction of occluded regions, we also use the corresponding non-middle decoded feature along with the warped features during frame synthesis as shown in \Eref{eqn:other_frames}. Similarly to the optical flow estimation stage, frames are synthesized in a bottom-up manner from the smallest to the full-scale resolution. 
\begin{equation}
    \{\hat{I}_{t_0}^l,\hat{I}_{t_1}^l \} = \calF_r^l\big(\{V_{t_0}^{l},V_{t_1}^{l}\} \oplus \{\hat{I}_{t_0}^{l+1},\hat{I}_{t_1}^{l+1}\} \big)
    \label{eqn:mid_frame}
\end{equation}
\begin{equation}
    \hat{I}^l_s = \calF_m^l\big(\calW\{V_{t_0}^{l}, \hat{f}^l_{s\rightarrow{t_0}}\} \oplus \calW\{V_{t_1}^{l}, \hat{f}^l_{s\rightarrow{t_1}}\} \oplus V_s^l \oplus \hat{I}_s^{l+1} \big)
    \label{eqn:other_frames}
\end{equation}
, where $s = \{t_{0}-N/2,\ldots,t_{0}-1,t_{0}+1,\ldots,t_{1}-1,t_{1}+1,\ldots,t_{1}+N/2\}$, $\calW$ denotes a warping layer and $\calF_m$ is a frame synthesis network for non-middle latent frames.

The proposed approach incorporates decoded features from both blurry inputs when estimating optical flows and predicting frames. This allows the frame synthesis network to exploit temporal and contextual information across inputs when interpolating and extrapolating latent frames. For instance, if the two consecutive inputs are substantially different in terms of blur sizes, our model leverages the less blurred input when predicting latent frames from the heavily blurred input, and hence, outputs a temporally smooth video with consistent visual quality. 
\paragraph{Network training.}
We train our network in an end-to-end manner by optimizing the estimated intermediate flows and predicted latent frames. For sharp frame reconstruction, we computed the $\ell1$ photometric loss between the predicted and ground truth frames. As our network predicts images at different scales, we used bilinear interpolation to downsample the ground truth frames to respective sizes. The weighted multi-scale photometric loss for reconstructing $N$ frames from two blurry frames is written as follows,
\begin{equation}
    \calL_{frame} = \sum_{n = 1}^N\sum_{l = 1}^{K}{w^l \cdot\big|I_n^l -\hat{I}_n^l\big|_1}
\end{equation}
, where $w^l$ is the frame loss weight coefficient at feature level $l$, $n$ is an index for the reconstructed frame sequence.

For optical flow training, we use endpoint error between the predicted flows and pseudo-ground truth flows. As mentioned earlier, we used pretrained FlowNet 2 \cite{IMKDB17} to compute the flows between the corresponding ground truth frames (from which the motion-blurred inputs are averaged) to guide the optical flow estimator. We predict a total of $2N-4$ optical flows when interpolating $N$ frames, and the weighted multi-scale endpoint error for supervising the estimated flows is computed as follows,
\begin{equation}
        \calL_{flow} = \sum_{m = 1}^{2N-4}\sum_{l = 1}^{K}{\hat{w}^l \cdot\big|f_m^l -\hat{f}_m^l\big|_2}
\end{equation}
, where $\hat{w}^l$ is a flow loss weight coefficient and $m$ is an index for estimated flows. The total training loss for interpolating and extrapolating $N$ number of sharp frames from two blurry input is given as a weighted sum of the two losses as shown below.
\begin{equation}
    \calL = \alpha_1\calL_{frame} + \alpha_2\calL_{flow}
\end{equation}

\section{Experiment}
\paragraph{Dataset.}
To train our network for the task at hand, we take advantage of two publicly available high speed video datasets to generate motion-blurred images. The GoPro high speed video dataset~\cite{Nah_2017_CVPR}, a benchmark for dynamic scene deblurring, provides 33 720P videos taken at 240\textit{fps}. We used 22 videos for training and generated motion-blurred images by averaging 7 consecutive frames. We also used the recently proposed Sony RX V high-frame rate video dataset \cite{Jin_2019_CVPR} which provides more than 60 1080P  videos captured at 250\textit{fps}. We used 40 videos during training and generated motion-blurred images by averaging 7 consecutive frames. To qualitatively and quantitatively analyze our approach on a diverse set of motion blurs, we choose 8 videos from each dataset (nonoverlapping with the training set) according to different blur sizes (small and large), blur types (static or dynamic) and complexity of the motion involved in the blurry video. We also provide generalization experiments on real motion-blurred videos from~\cite{su2017deep,Nah_2019_CVPR_Workshops_REDS}.

\paragraph{Implementation details.}
We implemented and trained our model in PyTorch \cite{paszke2019pytorch}. We used Adam \cite{KingmaB14} optimizer with parameters $\beta_1$, $\beta_2$ and \emph{weight decay} fixed to 0.9, 0.999 and $4e-4$, respectively. We trained our network using a mini-batch size of 4 image pairs by randomly cropping image patch sizes of $256\times256$. The pseudo-ground truth optical flows for supervising the predicted flows are computed on-the-fly during training. The loss weight coefficients are fixed to $w_6 = 0.32$, $w_5 = 0.08$, $w_4 = 0.04$, $w_3 = 0.02$, $w_2 = 0.01$ and $w_1 = 0.005$ from the lowest to the highest resolution, respectively, for both frames and flows. We trained our model for 120 epochs with initial learning rate fixed to $\lambda = 1e-4$ and gradually decayed by half at 60, 80 and 100 epochs. For the first 15 epochs, we only trained the optical flow estimator by setting $\alpha_1=0$ and $\alpha_2=1$ to facilitate feature decoding and flow estimation. For the rest of the epochs, we fixed $\alpha_1=1$ and $\alpha_2=1$. During inference, we interpolate and extrapolate frames by successively passing disjoint blurry frame pairs.

\subsection{Quantitative analysis}
In this section, we comprehensively analyze our work in connection to related works. Except for \cite{Jin_2018_CVPR} and our approach, other related methods fail to restore the first and the last few video frames. For fair evaluation purely based on the interpolated frames, we aligned the GT frames with the interpolated frames and discard the missing GT frames when evaluating such methods.
We perform motion-blurred video interpolation ($\times7$ slower video) and middle frame deblurring ($\times1$ video) comparisons on peak signal-to-noise ratio (PSNR) and structural similarity index measure (SSIM) metrics.

\begin{table}[!t]
\setlength{\tabcolsep}{4pt}
\renewcommand{\arraystretch}{1.0}
\begin{center}
\caption{Comparison with standard interpolation methods}
\label{tbl:interp}
\begin{tabular}{l|cc|cc}
\toprule
& \multicolumn{2}{c|}{GoPro} & \multicolumn{2}{c}{Sony RX V}  \\ \midrule
Method & PSNR & SSIM & PSNR & SSIM \\ \midrule
SepConv (Niklaus \etal) &26.977&0.769&26.181&0.716\\
SloMo (Jiang \etal) & 27.240 & 0.785 & 26.360 & 0.728\\
DAIN (Bao \etal) &27.220 & 0.783 & 26.410 & 0.731\\
Ours & \textbf{32.202} & \textbf{0.914} & \textbf{31.019} & \textbf{0.894}\\ \bottomrule
\end{tabular}
\end{center}

\end{table}

\begin{table}[!t]
\setlength{\tabcolsep}{5pt}
\renewcommand{\arraystretch}{0.8}
\begin{center}
\caption{Comparison with cascaded approaches}
\label{tbl:cascaded}
\begin{tabular}{l|cc|cc}
\toprule
& \multicolumn{2}{c|}{GoPro} & \multicolumn{2}{c}{Sony RX V}  \\ \midrule
Method & PSNR & SSIM & PSNR & SSIM \\ \midrule

DVD $\oplus$ DAIN & 25.650 & 0.722 & 27.885 & 0.791\\
DAIN $\oplus$ DVD & 28.885 & 0.843 & 28.157 & 0.797 \\ \midrule
DeepDeblur $\oplus$ DAIN & 28.154 & 0.831 & 27.192 & 0.782\\
DAIN $\oplus$ DeepDeblur & 28.176 & 0.829 & 27.195 & 0.778\\ \midrule
SRN $\oplus$ DAIN & 29.966 & 0.870 & 29.245 & 0.828\\ 
DAIN $\oplus$ SRN & 30.045 & 0.867& 29.074& 0.822\\ \midrule
Ours & \textbf{32.202} & \textbf{0.914} & \textbf{31.019} & \textbf{0.894}\\ \bottomrule
\end{tabular}
\end{center}
\vspace{-2.5mm}
\end{table}

\begin{figure*}[!t]
	\centering
 \begin{flushleft}
	\hspace{1.7cm} Blurry input \hspace{1.6cm} SRN + DAIN  \hspace{0.9cm}  Jin-Seq \hspace{1.1cm} Jin-SloMo \hspace{1.3cm}  Ours  \hspace{1.85cm} GT
 \end{flushleft}
	\includegraphics[width=1.0\textwidth]{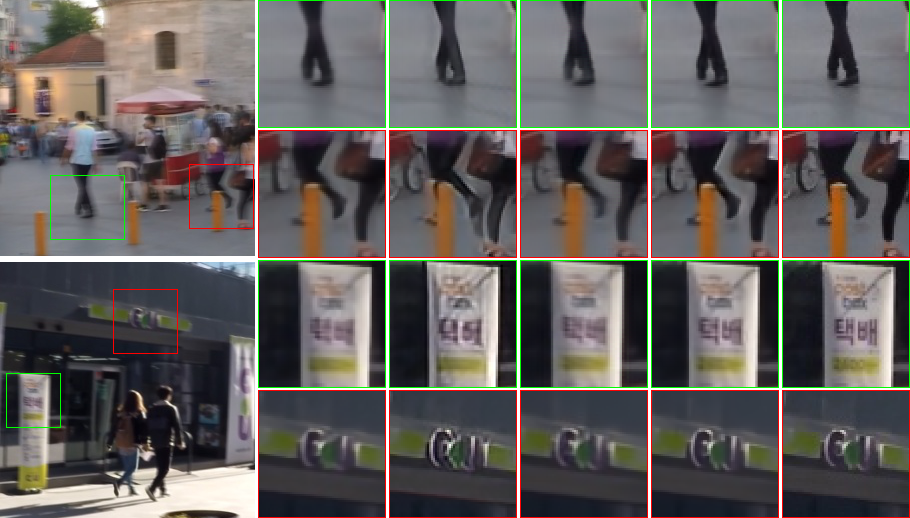}
	\caption{Qualitative analysis on interpolated frames. The $1^\mathrm{st}$ column shows blurry inputs from GoPro test set. The $2^\mathrm{nd}$ column depicts the outputs of cascaded approach (SRN \cite{tao2018srndeblur} + DAIN \cite{DAIN}). The $3^\mathrm{rd}$ column shows the outputs of Jin-Seq  \cite{Jin_2018_CVPR}. The $4^\mathrm{th}$ column shows frames interpolated by Jin-SloMo \cite{Jin_2019_CVPR} and the $5^\mathrm{th}$ column depicts the outputs of our network.}
	\label{fig:qual_compare}
\vspace{-2.5mm}
\end{figure*}

\paragraph{Cascaded approaches.}
One possible way to interpolate clean frames from given blurry inputs is to cascade interpolation and deblurring frameworks. To quantitatively analyze our method in comparison with such approaches, we experimented with state-of-the-art single image deblurring (DeepDeblur \cite{Nah_2017_CVPR}, SRN \cite{tao2018srndeblur}) and video deblurring (DVD \cite{su2017deep}) works cascaded with state-of-the-art interpolation methods (DAIN \cite{DAIN}, SloMo \cite{slomo}). As can be inferred from \Tref{tbl:cascaded}, our method performed consistently better than cascaded approaches. For instance, our approach outperforms the strongest baseline (SRN $\oplus$ DAIN) by a margin of 2.00 dB on average. This performance gain is mainly because cascaded approaches are prone to error propagation while our method directly interpolates clean frames from blurry inputs by estimating the motion within and across inputs. The effect of propagation of deblurring and interpolation artifacts can also be noticed from \Tref{tbl:mid_deblur}. Our method shows an average performance decrease of 0.71 dB on the interpolated videos ($\times 7$) compared to deblurred videos ($\times 1$) while SRN $\oplus$ DAIN shows an average performance decrease of 2.50 dB.
\paragraph{Comparison with previous works.} We compared our approach with works that restore sequence of latent frames from a single blurry input (Jin-Seq \cite{Jin_2018_CVPR}). Directly deploying such methods for motion-blurred video interpolation is not optimal since temporal ambiguity is a problem (see \Tref{tbl:previous_work}). To address this challenge, we applied our proposed flow-based rule during the inference stage by computing the necessary flows (between the restored frames) using pretrained FlowNet 2. This fix significantly improved performance by an average margin of 2.24 dB. While the sequence restoration can be achieved, contextual information between input frames is not exploited (as they are processed independently) leading to lower performances when compared to our approach.
We also analyzed our model in comparison with the recently proposed approach by Jin \etal (Jin-SloMo \cite{Jin_2019_CVPR}). As can be inferred from \Tref{tbl:previous_work}, our method outperforms Jin-SloMo by a margin of 1.82 dB and 1.50 dB on average on interpolated and deblurred videos, respectively. This is mainly because our method is relatively robust to large blurs while Jin-SloMo is limited to small motions (small blurs) as frames are deblurred and interpolated without taking pixel-level motion into consideration (see \Fref{fig:qual_compare}).
\begin{figure*}[!t]
\begin{center}
$B_1$ (Input) \hspace{1.25cm}  $B_2$  (Input) \hspace{0.95cm} $\hat{I}_1$ (Jin-SloMo) \hspace{1.15cm} $\hat{I}_1$ (Ours) \hspace{1.45cm} $\hat{I}_7$ (Ours) \hspace{1.4cm} $\hat{I}_{11}$ (Ours)
\renewcommand{\arraystretch}{0.6}
\setlength{\tabcolsep}{0.8pt}
\resizebox{1.0\linewidth}{!}{%
\begin{tabular}{cccccc}
        \includegraphics[width=0.1\linewidth]{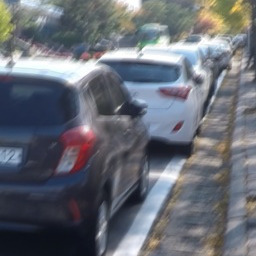} &
        \includegraphics[width=0.1\linewidth]{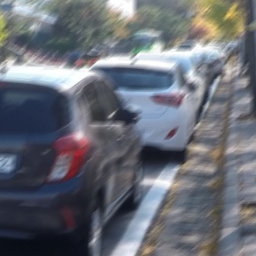} &
        \includegraphics[width=0.1\linewidth]{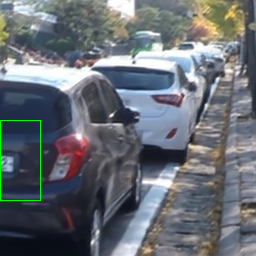} &
        \includegraphics[width=0.1\linewidth]{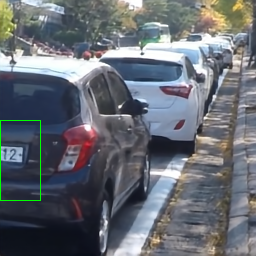} &
        \includegraphics[width=0.1\linewidth]{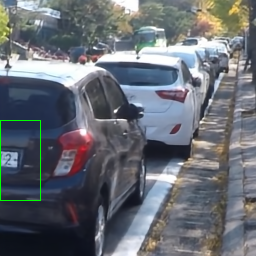} &
        \includegraphics[width=0.1\linewidth]{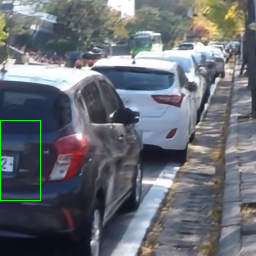}\\
        \includegraphics[width=0.1\linewidth]{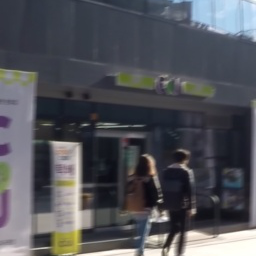} &
        \includegraphics[width=0.1\linewidth]{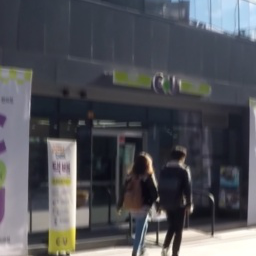} &
        \includegraphics[width=0.1\linewidth]{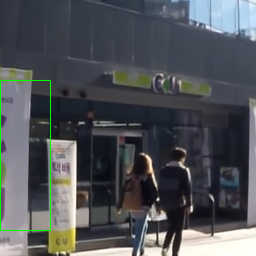} &
        \includegraphics[width=0.1\linewidth]{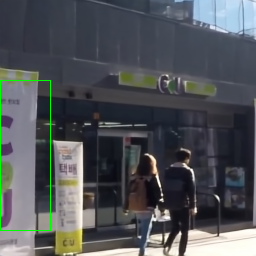} &
        \includegraphics[width=0.1\linewidth]{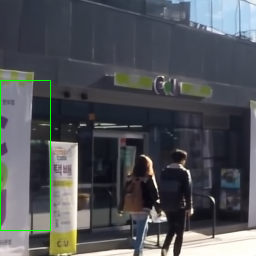} &
        \includegraphics[width=0.1\linewidth]{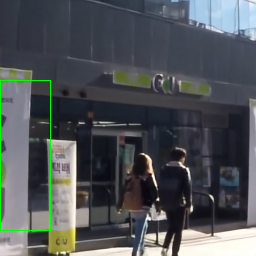}\\
\end{tabular}}
\end{center}
\caption{Qualitative analysis on extrapolated frames. Previous works \cite{Jin_2019_CVPR,shen2020blurry} ignore the first few latent frames in order not to deal with temporal ambiguity. In comparison, our approach outputs the entire latent frame sequence.}
\label{fig:extrapolation}
\vspace{-2mm}
\end{figure*}

\paragraph{Middle frame deblurring.}
Besides motion-blurred interpolation, we also analyzed the performance of our model for video deblurring \ie we evaluated the predicted reference (middle) latent frames. As can be inferred from \Tref{tbl:mid_deblur}, our approach performs competitively against state-of-the-art deblurring approaches. The slight performance loss can be attributed to the fact that deblurring works in general are trained with larger blurs (by averaging large number of frames, \eg DeepDeblur and SRN averaged 7-13 frames) while our work is trained on motion-blurred images generated by averaging 7 or 9 frames.
\begin{table}[!t]
\setlength{\tabcolsep}{5.5pt}
\renewcommand{\arraystretch}{0.8}
\begin{center}
\caption{Comparison with previous works}
\label{tbl:previous_work}
\begin{tabular}{l|cc|cc}
\toprule
& \multicolumn{2}{c|}{GoPro} & \multicolumn{2}{c}{Sony RX V}  \\ \midrule
Method & PSNR & SSIM & PSNR & SSIM \\ \midrule
Jin-Seq (2018)  & 26.848 & 0.785 & 25.785& 0.735 \\
Jin-Seq + \textit{flow fix} & 29.761 & 0.877 & 27.348 & 0.779 \\\midrule
Jin-SloMo (2019)   & 30.321 &0.878&29.267&0.816\\
Ours & \textbf{32.202} & \textbf{0.914} & \textbf{31.019} & \textbf{0.894}\\ \bottomrule
\end{tabular}
\end{center}
\end{table}

\begin{table}[!t]
\setlength{\tabcolsep}{4.0pt}
\renewcommand{\arraystretch}{0.8}
\begin{center}
\caption{Middle frame deblurring}
\label{tbl:mid_deblur}
\begin{tabular}{l|cc|cc}
\toprule
& \multicolumn{2}{c|}{GoPro} & \multicolumn{2}{c}{Sony RX V}  \\ \midrule
Method & PSNR & SSIM & PSNR & SSIM \\ \midrule
DVD (Su \etal) & 26.547 &0.742 & 28.937 & 0.805\\ 
DeepDeblur (Nah \etal) & 29.671 & 0.867 & 27.882 & 0.788\\
SRN (Tao \etal)& \textbf{33.382} & \textbf{0.931} & 30.827 & 0.851\\ \midrule
Jin-Seq (2018) & 31.442 & 0.906 & 29.752 &0.812 \\
Jin-SloMo (2019) & 31.318 & 0.900 & 30.325 & 0.829\\ 
Ours & 32.994 & 0.927 & \textbf{31.650} & \textbf{0.904}\\ \bottomrule
\end{tabular}
\end{center}
\vspace{-2.5mm}
\end{table}
\subsection{Qualitative analysis}

\paragraph{Interpolated frames.}  We qualitatively compared our approach with related works on the quality of the interpolated frames. As can be seen from \Fref{fig:qual_compare}, our approach is relatively robust to heavily blurred inputs and interpolates visually sharper images with clearer contents compared to other related methods \cite{tao2018srndeblur,DAIN,Jin_2018_CVPR,Jin_2019_CVPR}.

\begin{figure*}[!t]
\begin{center}
\renewcommand{\arraystretch}{0.6}
\setlength{\tabcolsep}{0.8pt}
\resizebox{1.0\linewidth}{!}{%
\footnotesize
\begin{tabular}{ccccccc}
        \raisebox{2.1\normalbaselineskip}[0pt][0pt]{\scriptsize{{\rotatebox[origin=c]{90}{Blurry input}}}}&
        \includegraphics[width=0.1\linewidth]{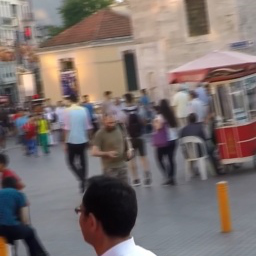} &
        \includegraphics[width=0.1\linewidth]{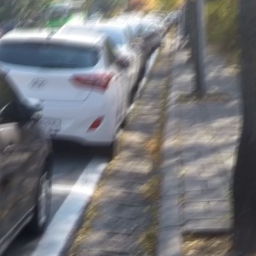} &
        \includegraphics[width=0.1\linewidth]{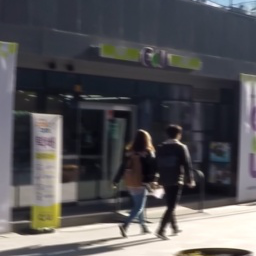} &
        \includegraphics[width=0.1\linewidth]{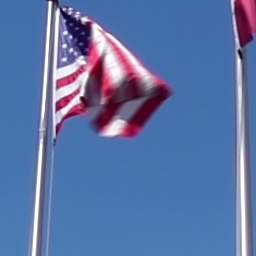} &
        \includegraphics[width=0.1\linewidth]{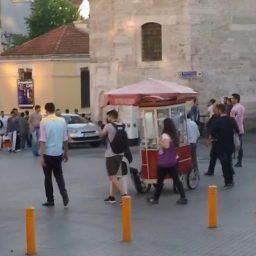} &
        \includegraphics[width=0.1\linewidth]{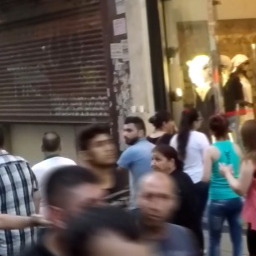} \\
        \raisebox{2.1\normalbaselineskip}[0pt][0pt]{\scriptsize{{\rotatebox[origin=c]{90}{Ours}}}}&
        \includegraphics[width=0.1\linewidth]{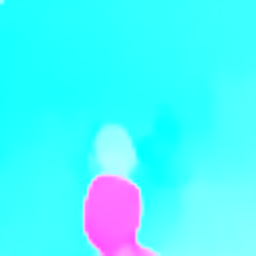}&
        \includegraphics[width=0.1\linewidth]{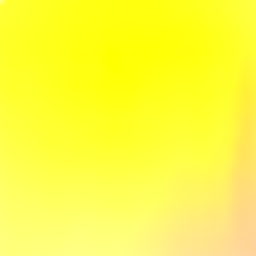}&
        \includegraphics[width=0.1\linewidth]{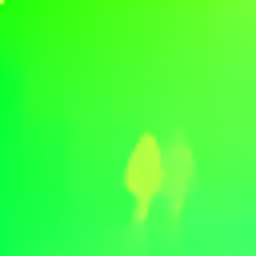} &
         \includegraphics[width=0.1\linewidth]{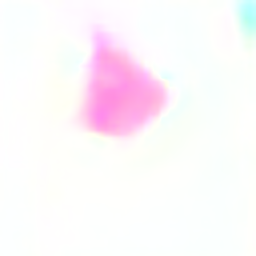} &
        \includegraphics[width=0.1\linewidth]{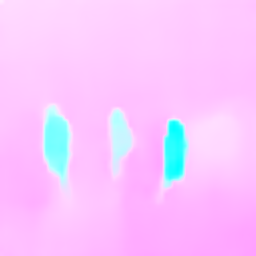} &
        \includegraphics[width=0.1\linewidth]{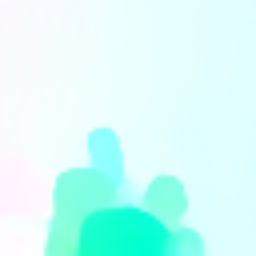}\\
                \raisebox{2.1\normalbaselineskip}[0pt][0pt]{\scriptsize{{\rotatebox[origin=c]{90}{p-GT}}}}&
        \includegraphics[width=0.1\linewidth]{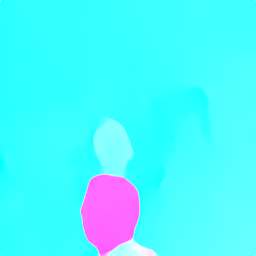}&
        \includegraphics[width=0.1\linewidth]{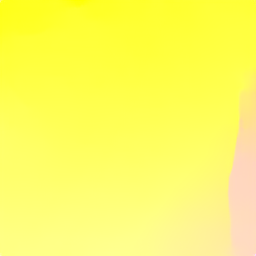}&
        \includegraphics[width=0.1\linewidth]{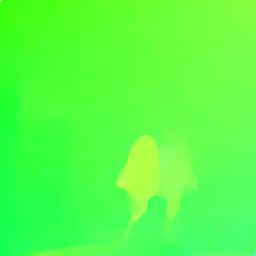} &
         \includegraphics[width=0.1\linewidth]{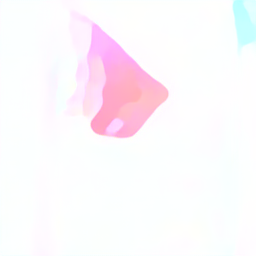} &
        \includegraphics[width=0.1\linewidth]{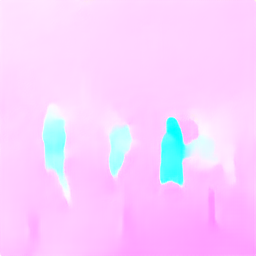} &
        \includegraphics[width=0.1\linewidth]{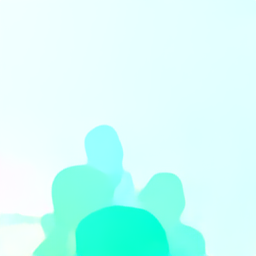}
\end{tabular}}
\end{center}
\caption{Qualitative analysis on estimated optical flows. The second row depicts optical flows estimated by our model from blurry inputs and the third row shows the corresponding p-GT flows from sharp latent frames.}
\label{fig:qual_flow}
\end{figure*}

\paragraph{Extrapolated frames.} Previous works \cite{Jin_2019_CVPR,shen2020blurry} implicitly follow a \emph{deblurring} $\rightarrow$ \emph{interpolation} pipeline, and hence can only interpolate frames between the reference latent frames. Our approach, on the other hand, not only interpolates intermediate frames but also extrapolates the latent frames underlying to the left and right side of the reference latent frames. As shown in \Fref{fig:extrapolation}, the $1^\mathrm{st}$ frame interpolated by Jin-SloMo is aligned with the $11^{\mathrm{th}}$ frame predicted by our approach. Their approach ignores the first 10 latent frames so as not to deal with potential temporal ambiguity. By contrast, our work reconstructs the entire latent frame sequence in a temporally coherent manner.
\paragraph{Estimated optical flows.} We qualitatively analyzed the intermediate optical flows estimated by our approach in comparison with pseudo-ground truth (p-GT) flows predicted from the corresponding sharp latent frames using pretrained FlowNet 2. As can be inferred from \Fref{fig:qual_flow}, our network estimates accurate optical flows from decoded features of blurry inputs for different blur types involving dynamic motion of multiple objects in a close to static or moving scene. This further explains the quantitative performance of our approach for motion-blurred video interpolation in the previous section as estimating correct pixel-level motion is crucial for accurate frame interpolation.


\section{Ablation studies}
\paragraph{Optical flow estimation. } To examine the importance of optical flow estimation, we directly regressed latent frames from the decoded features (without estimating flow) using only the frame synthesis network \ie $ \{\hat{I}_{t_{0}-N/2}^l,\ldots,\hat{I}_{t_{1}+N/2}^l \} = \calF^l\big(V_{t_{0}-N/2}^{l},\ldots,V_{t_{1}+N/2}^{l}\big)$. 
The results on motion-blurred video interpolation ($\times 7$) are summarized in \Tref{tbl:ablation}. We experimentally observed that $\calL_{frame}$ is a strong enough constraint to guide the STNs and motion decoders to decode features of each blurry input in a temporally ordered manner (without random shuffling), yet, correct temporal direction can not be guaranteed. To ensure temporal coherence, we ordered the predicted frames using the proposed flow-based rule. This post-processing step improves performance by 0.78 dB. Even with the flow fix, however, directly predicting frames without motion estimation causes a performance decrease of 1.92 dB on average compared to our full model. This highlights that estimating optical flows is not only useful to address temporal ambiguity but also important to warp decoded features for sharp reconstruction of latent frames.
\vspace{-2.5mm}
\paragraph{Feature decoding.} Motion decoders ($\calD_m$) decode non-middle latent features with respect to reference latent features by refining the STN transformed features (see \Eref{eqn:nonmid_decoding1} and \Eref{eqn:nonmid_decoding2}). Training our network without motion decoders, \ie decoding features only via STN transformation, results in a subpar performance as shown \Tref{tbl:ablation}. This is mainly because local motions that are apparent in the high speed videos can not be effectively captured only using STNs. In principle, both global and local motions can be implicitly learnt by guiding motion decoders (without the need to explicitly model global motions with STNs) via optical flow supervision as CNNs have been shown to be effective in motion estimation tasks. This is also empirically evident as a network trained with only motion decoders results in a good performance (see \Tref{tbl:ablation}). However, incorporating STNs to learn global motions also proved to give a significant performance boost of 0.89 dB.

\begin{table}[!t]
\setlength{\tabcolsep}{5pt}
\renewcommand{\arraystretch}{0.8}
\begin{center}
\caption{Ablation studies}
\label{tbl:ablation}

\begin{tabular}{ccc|cc|cc}
\toprule
 \multicolumn{3}{c|}{}& \multicolumn{2}{c|}{GoPro} & \multicolumn{2}{c}{Sony RX V}  \\ \midrule
STN & $\calD_m$ & Flow & PSNR & SSIM & PSNR & SSIM \\ \midrule
\cmark & \cmark & \xmark & 29.509 & 0.836 & 28.316 & 0.805\\
\cmark & \cmark &  \textit{flow fix} & 30.219 & 0.870 & 29.163 & 0.812 \\ \midrule
\cmark & \xmark &  \cmark & 28.789 & 0.855 & 27.467 & 0.798\\
\xmark & \cmark &  \cmark & 31.317 & 0.893 & 30.125 & 0.857\\
\cmark & \cmark &  \cmark & \textbf{32.202} & \textbf{0.914} & \textbf{31.019} & \textbf{0.894}\\ \bottomrule
\end{tabular}
\end{center}
\vspace{-2.5mm}
\end{table}

\section{Conclusion}
In this work, we tackle the problem of multi-frame interpolation and extrapolation from a given motion-blurred video. We adopt a motion-based approach to predict frames in a temporally coherent manner without ambiguity. As a result, our method can interpolate, extrapolate and recover high quality frames in a single pass. Our method is extensively analyzed in comparison with existing approaches. We also experimented with the applicability of our approach on related tasks such as video deblurring and flow estimation.

\paragraph{Acknowledgement}
This work was supported by NAVER LABS Corporation [SSIM: Semantic \& scalable indoor mapping].

\bibliography{egbib}
\end{document}

%% file: def.tex
\newcommand{\calC}{{\mathcal{C}}}
\newcommand{\calD}{{\mathcal{D}}}
\newcommand{\calE}{{\mathcal{E}}}
\newcommand{\calF}{{\mathcal{F}}}

\newcommand{\calL}{{\mathcal{L}}}

\newcommand{\calO}{{\mathcal{O}}}

\newcommand{\calW}{{\mathcal{W}}}

\newcommand{\be}{\begin{eqnarray}}
\newcommand{\ee}{\end{eqnarray}}
\newcommand{\bee}{\begin{eqnarray*}}
\newcommand{\eee}{\end{eqnarray*}}

\newcommand{\matrixb}{\left[ \begin{array}}
\newcommand{\matrixe}{\end{array} \right]}   

\newcommand{\Tref}[1]{Table~\ref{#1}}
\newcommand{\Eref}[1]{Eq.~(\ref{#1})}
\newcommand{\Fref}[1]{Fig.~\ref{#1}}

\newcommand{\Cref}[1]{Chap.~\ref{#1}}

\def\eg{\emph{e.g. }}

\def\etal{\emph{et al. }}
\def\ie{\emph{i.e. }}